\documentclass{ametsocV6.1_arxiv}

\newcommand{\arrtime}{{\bm{\tau}}}
\newcommand{\arrtimegen}{{\bm{\tau}^{\bm{g}}}}
\newcommand{\arrtimeRV}{{\bm{T}}}
\newcommand{\hgt}{{\bm{h}}}
\newcommand{\hgtRV}{{\bm{H}}}
\newcommand{\arrtimemeas}{{\bm{\bar{\tau}}}}
\newcommand{\arrtimemeasRV}{{\bm{\bar{T}}}}
\newcommand{\cond}{{P_{\bm{T}|\bm{\bar{T}},\bm{H}}}}
\newcommand{\condgen}{{P_{\bm{T}|\bm{\bar{T}},\bm{H}}^{\bm{g}}}}
\newcommand{\joint}{{P_{\bm{T,\bar{T},H}}}}
\newcommand{\latent}{\bm{z}}
\newcommand{\latentRV}{\bm{Z}}
\newcommand{\latentdist}{P_{\bm{Z}}}

\usepackage{makecell}

\usepackage{soul}

\title{Generative Algorithms for Wildfire Progression Reconstruction from Multi-Modal Satellite Active Fire Measurements and Terrain Height}

\authors{Bryan Shaddy,\aff{a}\correspondingauthor{Bryan Shaddy, bshaddy@usc.edu \\ \emph{This Work has not yet been peer-reviewed and is provided by the contributing Author(s) as a means to ensure timely dissemination of scholarly and technical Work on a noncommercial basis. Copyright and all rights therein are maintained by the Author(s) or by other copyright owners. It is understood that all persons copying this information will adhere to the terms and constraints invoked by each Author's copyright. This Work may not be reposted without explicit permission of the copyright owner.}} 
Brianna Binder,\aff{a} 
Agnimitra Dasgupta,\aff{a}
Haitong Qin,\aff{b}
James Haley,\aff{c} 
Angel Farguell,\aff{d} 
Kyle Hilburn,\aff{c} 
Derek V. Mallia,\aff{e} 
Adam Kochanski,\aff{d} 
Jan Mandel,\aff{f} 
and Assad Oberai\aff{a}
}

\affiliation{\aff{a}{Department of Aerospace and Mechanical Engineering, University of Southern California, Los Angeles, California} \\
\aff{b}{Department of Mathematics, University of Southern California, Los Angeles, California}\\ 
\aff{c}{Cooperative Institute for Research in the Atmosphere, Colorado State University, Fort Collins, Colorado}\\ 
\aff{d}{Department of Meteorology and Climate Science, San Jose State University, San Jose, California}\\ 
\aff{e}{Department of Atmospheric Sciences, University of Utah, Salt Lake City, Utah}\\
\aff{f}{Department of Mathematical and Statistical Sciences, University of Colorado Denver, Denver, Colorado}
}

\abstract{Increasing wildfire occurrence has spurred growing interest in wildfire spread prediction. However, even the most complex wildfire models diverge from observed progression during multi-day simulations, motivating need for data assimilation. A useful approach to assimilating measurement data into complex coupled atmosphere-wildfire models is to estimate wildfire progression from measurements and use this progression to develop a matching atmospheric state. In this study, an approach is developed for estimating fire progression from VIIRS active fire measurements, GOES-derived ignition times, and terrain height data. A conditional Generative Adversarial Network is trained with simulations of historic wildfires from the atmosphere-wildfire model WRF-SFIRE, thus allowing incorporation of WRF-SFIRE physics into estimates. Fire progression is succinctly represented by fire arrival time, and measurements for training are obtained by applying an approximate observation operator to WRF-SFIRE solutions, eliminating need for satellite data during training. The model is trained on tuples of fire arrival times, measurements, and terrain, and once trained leverages measurements of real fires and corresponding terrain data to generate samples of fire arrival times. The approach is validated on five Pacific US wildfires, with results compared against high-resolution perimeters measured via aircraft, finding an average S\o rensen-Dice coefficient of 0.81. The influence of terrain height on the arrival time inference is also evaluated and it is observed that terrain has minimal influence when the inference is conditioned on satellite measurements.}

\begin{document}

\maketitle

%
%
%
\statement

We present an approach to estimate initial wildfire progression using satellite measurements and terrain data in a way that is consistent with simulations of historic wildfires. This approach may be used to initialize forecasts of future wildfires thereby enabling the accurate prediction of these events. The work developed here follows a probabilistic framework, providing an ensemble of initial wildfire progression estimates conditioned on measurements and terrain, thus allowing for the initialization of an ensemble of wildfire simulations which would in turn enable the quantification of uncertainty. Initial wildfire progression estimates are compared to high-resolution perimeters measured via aircraft, yielding an average S\o rensen-Dice coefficient of 0.81, indicating that predicted perimeters compare well to measured perimeters.

\section{Introduction}
Climate trends and variability including drier summers, wetter winters, and more heat have led to an increased rate of wildfire occurrence during the end of the 20th and beginning of the 21st centuries \citep{dennison2014large}. Predictions for the remainder of this century indicate larger wildfires with increased intensity, accompanied by a lengthening of wildfire season duration for many regions globally \citep{westerling2006warming,flannigan2013global}. Negative impacts associated with increased wildfire presence are compounded by growing numbers of residential homes on the wildland-urban interface in the United States, with current estimates putting the number at around 50 million and increasing by approximately 350,000 houses per year, leading to greater potential for communities to be directly affected by wildfires \citep{burke2021changing}. Due in part to the increased proximity to wildfires, economic impact stemming from wildfires has also increased, with damages from the 2018 California wildfires estimated at nearly $\$150$ billion \citep{wang2021economic}.This estimate includes capital losses, health costs, and indirect losses. Further, initial estimates of the economic impact from the Los Angeles wildfires during the first two weeks of 2025 alone are around $\$250$ billion \citep{Danielle_2025}. 

Wildfire spread is influenced by the interaction of fire with fuels, topography, and weather conditions, a well studied set of relations formally quantified in the 1970's by Rothermel's wildfire rate of spread model, which has served as the basis for many wildfire behavior models \citep{rothermel1972mathematical,albini1976estimating}. With the increased size and intensity of modern wildfires, interactions with the atmosphere are increasing in complexity and becoming an even more critical aspect of predicting fire behavior \citep{lareau2016environmental,lareau2017mean,pastor2003mathematical}. To capture interactions between wildfires and the atmosphere, in addition to topography and fuel, state-of-the-art models such as WRF-SFIRE consider two-way atmosphere-wildfire coupling, where heat released to the atmosphere by a wildfire is used to modify local meteorological conditions, which are then fed back into the spread of the fire \citep{bakhshaii2019review,mandel2011coupled}. However, even forecasts from these complex models which aim to capture all critical aspects of wildfire spread ultimately diverge from the true state of a wildfire over the course of a one- to two-day simulation. This reduction in accuracy underscores the need for data assimilation, which integrates measurements of ongoing wildfires to enhance prediction quality.

Two approaches have proven useful for performing data assimilation in coupled atmosphere-wildfire models. The first involves using a series of measurements for a wildfire during its initial spread to estimate the fire and atmosphere states, enabling subsequent forecasts; this can be thought of as initial state estimation. The second approach involves using current predictions of the wildfire and atmosphere states along with a newly obtained set of measurements to determine updated states; this can be thought of as sequential state estimation. In either case it is critical to recognize that both wildfire and atmosphere variables must be modified in a self-consistent manner. To accomplish this, \cite{mandel2012assimilation} recognized that if the precise spread of a wildfire can be determined it may be prescribed within a coupled atmosphere-wildfire model to provide heat and mass fluxes at the right place and time to generate a corresponding atmospheric state. This process is referred to as spin-up, and simplifies the state estimation problem to one concerned solely with estimating fire progression. It is convenient to represent wildfire spread using the fire arrival time, which is the time from ignition it takes for the fire front to reach a given location. This provides a succinct representation of fire progression in a single field which can be represented on a spatial grid by a vector \citep{mandel2012assimilation,mandel2014recent}. Therefore, for either problem, we are interested in estimating fire arrival times for some interval (either from ignition to current time or the previous DA step to the current DA step) and using this to develop a corresponding atmospheric state for this interval, allowing subsequent forecasting. In this paper we address the initial state estimation problem. 

Common systems for monitoring wildfire spread include satellite-based sensors deployed on polar orbiting and geostationary satellites. These satellite systems utilize thermal imaging to provide active fire (AF) detections, which indicate the time and location an active fire was identified for a given area. The Visible Infrared Imaging Radiometer Suite (VIIRS), which is carried on polar-orbiting satellites including Suomi NPP, NOAA-20, and NOAA-21, provides AF detections with a spatial resolution of 375 m roughly 2-4 times per day at mid-latitudes \citep{schroeder2024collection}. The geostationary GOES satellite series, serving operationally as GOES-East and GOES-West, provides similar data through the Fire / Hot Spot Characterization product, producing measurements every 5 minutes, though at a coarser spatial resolution of 2-4 km \citep{schmidt2010goes}. AF detections from these satellite systems are useful for providing information about wildfire progression, however, the data provided are at spatial and temporal scales that are orders of magnitude coarser than those of modern wildfire spread models, complicating their use for initializing these models. Data from these systems also tends to be sparse and noisy due to artifacts caused by sun glint, clouds, topography, and in some cases hot smoke or embers. However, even with these limitations, AF data serves as a useful source of measurements for performing data assimilation, where it may be used as a basis for inferring a smooth and finely-sampled version of the fire arrival time. 

Limitations of previous approaches for estimating fire arrival times from AF detections include reliance on deterministic methods, use of measurements exclusively from polar-orbiting satellite systems, inclusion of little-to-no physics, and the inability to incorporate additional types of data such as terrain height or fuel type. In \cite{farguell2021machine}, a method based on the support vector machine (SVM) was applied to estimate fire arrival times using satellite AF detections and clear ground pixels. In this approach, SVM is used to find an optimal separation between AF detections and clear ground pixels in space and time to generate an approximate fire arrival time map. While this approach has provided consistent results, it must be trained for each new set of measurements and does not allow for uncertainty quantification, which is useful when dealing with error-prone data as provided by satellite measurements. Further, this approach does not incorporate wildfire spread physics; nor does it consider static data such as terrain when making its predictions. Inspired by the SVM approach, \cite{shaddy2024generative} proposed the use of a conditional Wasserstein Generative Adversarial Network trained using WRF-SFIRE solutions to probabilistically reconstruct fire arrival times from AF measurements. While the cWGAN-based approach allowed for the addition of uncertainty quantification during inference and the inclusion of physics through training data, only data from the VIIRS system was considered for fire arrival time reconstruction. Additionally, the approach did not include conditioning on additional fields like terrain or fuel. 

The present work aims to expand upon the cWGAN-based approach of \cite{shaddy2024generative} in several significant ways. We significantly increase the complexity of wildfire spread solutions used for training by using WRF-SFIRE simulation results for real wildfires occurring in CONUS during 2023, thus allowing more complex and realistic wildfire spread physics to be incorporated into fire arrival time estimates. This is in contrast to \cite{shaddy2024generative}, where the training data comprised arrival times for synthetically created wildfires occurring on flat terrain with a uniform fuel type and simplified atmospheric conditions. Secondly, we consider the influence of terrain during fire arrival time inference by including terrain as an additional conditioning variable. Thirdly, we take advantage of AF data from GOES by including it to estimate the ignition time during fire arrival time inference. 

To achieve these goals, our approach begins with the generation of wildfire spread solutions using WRF-SFIRE, where we simulate 48 hours of spread for named wildfires occurring in CONUS during 2023, which provide the target fire arrival times along with their corresponding terrain height maps. We then apply an approximate observation operator to the fire arrival times to simulate measurements from the VIIRS system with ignition times derived from GOES. The arrival times, terrain, and measurement are treated as random variables, and the generated 3-tuples of these variables represent samples from a joint distribution. These tuples are used to train a conditional Wasserstein Generative Adversarial Network (cWGAN) for sampling from the conditional distribution of arrival times conditioned on measurements and terrain. The cWGAN is trained using an adversarial loss, allowing for the inclusion of realistic wildfire physics through the WRF-SFIRE solutions used for training. Once trained, the cWGAN is used to draw samples from the desired conditional distribution of the inferred field (fire arrival times), given the conditioning variables (AF measurements and terrain). The samples thus obtained are used to compute statistics of interest including the mean and standard deviation, providing a best guess estimate and a measure of uncertainty. Our approach is applied retrospectively to wildfires occurring on the US West Coast, and results are validated using high-resolution fire perimeters measured using aircraft-mounted infrared sensors. Finally, analysis of the influence of terrain conditioning on arrival time reconstruction is presented and discussed. 

The remainder of the manuscript is structured as follows. In Section \ref{problem_description}, we describe and formulate the problem of inferring fire arrival times from satellite measurements and terrain height data. Thereafter, in Section \ref{training_data}, we discuss the wildfire spread solutions generated using WRF-SFIRE and the process to transform them into the training data used here. Following this, in Section \ref{cWGAN} we describe the cWGAN model utilized and provide details of training. In Section \ref{results}, results are then presented and validated for five Pacific US wildfires, and the influence of terrain conditioning is further examined. Lastly, in Section \ref{conclusion}, we provide concluding remarks and look towards possible future developments.

\section{Problem Description} \label{problem_description}
Let the vector of fire arrival times be denoted by the random vector $\arrtimeRV$ with the probability density $P_\arrtimeRV$. Samples $\arrtime$ of $\arrtimeRV$ consist of $N$ components $\tau_{i}$ and therefore $\arrtime \in \Omega_{\arrtime} \equiv \mathbb{R}^{N}$. The components of $\arrtime$ are measured on a 12.8 km $\times$ 12.8 km domain with a resolution of 25 m and there are $ N = 512 \times  512$ such components. The terrain height vector is described on the same grid and is represented by the random vector $\hgtRV$ with a probability density $P_{\hgtRV}$ and samples $\hgt \in \Omega_{\hgt} \equiv \mathbb{R}^{N}$.

An observation operator $M$ defined by the mapping $M: \Omega_{\arrtime} \rightarrow \Omega_{\arrtimemeas}$ may be applied to fire arrival times $\arrtime$ to generate coarse, sparse and noisy measurements of arrival time $M({\arrtime}) = \arrtimemeas$, with their own probability density $P_{\arrtimemeasRV}$. The measured arrival times are defined on the same grid, and therefore $\arrtimemeas \in \Omega_{\arrtimemeas} \equiv \mathbb{R}^{N}$. Here the mapping $M$ is defined through a set of steps described in Section \ref{training_data}\ref{meas_steps} that approximate the measurements generated by satellite active fire products.

We are interested in solving for the inverse mapping which transforms measurements $\arrtimemeas$ to fire arrival times $\arrtime$. Further, we would like to condition this inference on terrain height $\hgt$. We note that, unlike the forward mapping, the inverse mapping is ill-posed because a single measurement-terrain pair $(\arrtimemeas,\hgt)$ can correspond to many different fire arrival times $\arrtime$.

To address the ill-posed nature of the inverse map we propose a probabilistic approach wherein we quantify the conditional distribution $\cond$ such that for a given measurement-terrain pair $(\arrtimemeas,\hgt)$ we are able to draw samples of $\arrtime$ from this conditional distribution. To accomplish this, we use a cWGAN trained on samples of $(\arrtime,\arrtimemeas,\hgt)$ drawn from the joint distribution $\joint$. Once trained, for a given $(\arrtimemeas,\hgt)$ pair the generator of the cWGAN produces samples of $\arrtime$ from the learned conditional distribution $\condgen \approx \cond$. These samples are used to compute statistics such as the mean and standard deviation of the arrival time.

To generate training data, first samples of terrain height $\hgt$ are drawn from the marginal distribution $P_\hgtRV$. Following this, WRF-SFIRE is employed to draw corresponding samples of fire arrival times $\arrtime$ from the conditional distribution $P_{\arrtimeRV|\hgtRV}$ by running wildfire progression simulations for a given $\hgt$. The measurement operator $M$ is then applied to samples of $\arrtime$ to provide corresponding samples $\arrtimemeas$. Thus, tuples of $(\arrtime,\arrtimemeas,\hgt)$ from the joint distribution $\joint$ are obtained. With samples from the joint distribution obtained, the cWGAN may be trained to sample from any corresponding conditional distribution, where here we are interested in sampling from $\cond$. The following section includes a detailed discussion of the training data generation process. 

\section{Training Data} \label{training_data}
To generate training data consisting of tuples of fire arrival times $\arrtime$, measurements $\arrtimemeas$, and terrain height $\hgt$, we employ the fully coupled atmosphere-wildfire model WRF-SFIRE, which couples the Weather Research and Forecasting model with a fire spread parameterization (SFIRE). SFIRE is based on the level-set method and uses the semi-empirical Rothermel rate of spread model to propagate fire growth \citep{mandel2011coupled,rothermel1972mathematical}. WRF-SFIRE simulations were run utilizing the wrfxpy system for automated simulation initiation, which configures the simulation domain, collects and processes data for initial conditions and boundary conditions, and initiates the simulation. The following sections will outline the use of the WRF-SFIRE model to produce solutions for training, followed by the construction of the corresponding measurements.

\subsection{WRF-SFIRE simulations}
For training, results from 140 WRF-SFIRE simulations of named wildfires occurring in CONUS during 2023 were utilized, with an additional 12 WRF-SFIRE simulations reserved for validation / hyperparameter tuning. To initialize the WRF-SFIRE simulations, ignition locations and times were derived from GOES and VIIRS active fire detections, delivered by NOAA's Next Generation Fire System (NGFS). The algorithm used by the NGFS clusters detections into incidents which are then associated with reported fires cataloged by the National Interagency Fire Center (NIFC). The ignition time is estimated to be the time of the first detection belonging to a cluster of detections associated with a fire event. The ignition location is estimated from the average location of all VIIRS detections found within the footprint of the initial GOES detection. If no VIIRS detections are observed within the first four hours after the initial GOES detection, an average of the GOES detection locations is used.

Simulations were run using a domain of size 30 km x 30 km, with an atmospheric grid of size 30 x 30 with 1 km resolution. The atmospheric grid additionally contained 40 vertical levels, with a pressure top of roughly 5000 Pa (approximately 20 km). The wildfire state was tracked on a grid of size 1200 x 1200 with 25 m resolution, using a subdivision of the atmospheric mesh. Domains were constructed with ignition locations approximately in the center. Initial and boundary conditions were obtained from the North American Mesoscale (NAM) Forecast System Grid 227, which was available at 3 hour intervals with 5 km resolution. NAM provides data for a variety of meteorological variables such as wind, temperature, pressure, and humidity, and was also used to provide boundary conditions for the fuel moisture model. Fuel category data came from the 2019 LandFire 13 Anderson Fire Behavior Fuel Models (FBFM13) dataset and terrain height data was obtained from the USGS National Elevation Dataset (NED). With the simulations initialized, 48 hours of fire spread was simulated, providing the fire arrival time maps $\arrtime$ of interest.

Once fire arrival time maps $\arrtime$ and corresponding terrain height maps $\hgt$ are obtained, arrival time values are adjusted to ensure ignition occurs at time zero, after which data augmentation is performed to the tuples $(\arrtime,\hgt)$ in the following steps:
 \begin{enumerate}
     \item Rotate randomly between 0 and 360 degrees.
     \item Translate randomly in a box of size 1 km x 1 km.
     \item Crop to a size of 512 x 512 pixels (12.8 km x 12.8 km).
 \end{enumerate}

The above data augmentation steps were performed 25 times per each WRF-SFIRE fire arrival time solution, following which the observation operator was applied to each augmented solution 5 times, creating 5 unique measurements per augmented target fire arrival time and terrain pair. In total this leads to 17,500 tuples $(\arrtime^{(i)},\arrtimemeas^{(i)},\hgt^{(i)})$ for training and an additional 1,500 tuples for validation. Additional details about the observation operator are provided in the following section. 

\subsection{Measurement construction} \label{meas_steps}
With fire arrival times $\arrtime^{(i)}$ obtained, an approximation of the observation operator $M$ is then applied to generate corresponding measurements $\arrtimemeas^{(i)}$ similar to what is obtained by detections from the 375 m Level-2 VIIRS AF product, with ignition times obtained from GOES Fire / Hot Spot Characterization data. The measurement operator is approximated and applied in the following set of steps:

\begin{enumerate}
    \item Coarsen $\arrtime^{(i)}$ to a resolution of 375 m using a box kernel. This accounts for the coarser resolution of VIIRS data.
    \item Create four copies of the coarsened $\arrtime^{(i)}$ and denote them as $\arrtime^{(i)}_j, j=1,...,4$. This accounts for presence of multiple VIIRS measurements during the initial progression of a fire.
    \item With a probability of $1/2$ eliminate or retain components of $\arrtime^{(i)}_j$. This accounts for independent noise between individual VIIRS measurements.
    \item Sample 4 measurement times $t_j, j=1,...,4$ from a uniform probability distribution $U(2,max(\arrtime^{(i)})-0.1)$ and sort them in ascending order. This accounts for randomness in the time of VIIRS measurements.
    \item For each measurement time $t_j$ generate a time interval $(t_j-\delta, t_j)$, where $\delta$ is sampled from $U(6,12)$. If $t_j-\delta < 0$ set it to $0$. 
    \item For $j = 1, \cdots,4$ set fire arrival time values in $\arrtime^{(i)}_j$ falling within the associated interval $(t_j-\delta, t_j)$ to $t_j$. Set arrival time values outside of this interval to a background value. This allows pixels ignited prior to a VIIRS measurement which may still be burning during a subsequent measurement to be captured. 
    \item Combine measurements $\arrtime^{(i)}_j$ into a single measurements by taking $\arrtime^{(i)} = min_j(\arrtime^{(i)}_j)$ at each pixel.
    \item Choose an ignition time error $\delta T$ from $U(0,2)$ and subtract it from arrival time values in $\arrtime^{(i)}$. This accounts for error in ignition time estimates obtained from GOES data.
    \item Randomly eliminate two 3 km x 3 km patches from $\arrtime^{(i)}$ and set eliminated pixels to background value. This accounts for persistent obstructions occurring across all VIIRS measurements. 
    \item Set all background pixel values to 48 h.
    \item Upsample to original resolution of 25 m and let the resulting measurement be denoted by $\arrtimemeas^{(i)}$.
\end{enumerate}

\begin{figure}
    \centering
    \includegraphics[width=\linewidth]{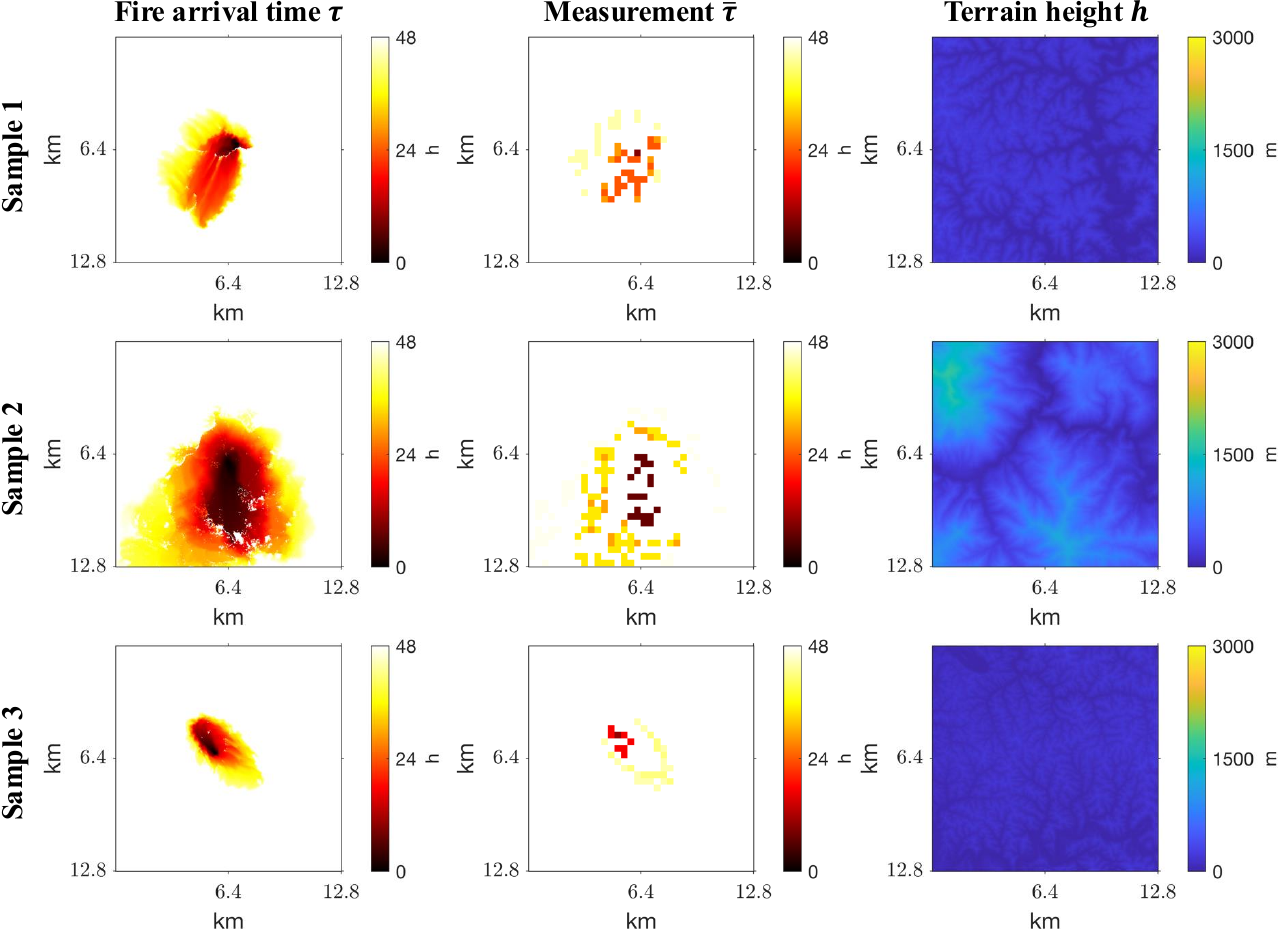}
    \caption{Training data samples, with target fire arrival time solutions $\arrtime$ in the first column, measurements $\arrtimemeas$ in the second column, and terrain height $\hgt$ in the third column.}
    \label{fig:training_data}
\end{figure}

With these steps, artificial measurements $\arrtimemeas^{(i)}$ are generated in a way that approximates a sequence of VIIRS measurements from multiple satellite overpasses, with ignition times derived from GOES. By simulating critical characteristics of active fire data, resulting artificial measurements contain the same sparse, coarse, and noisy patterns as observed for true measurements. Fig.~\ref{fig:training_data} shows sample tuples from the training data set. Following this and prior to training, fire arrival times $\arrtime^{(i)}$ and measurements $\arrtimemeas^{(i)}$ are normalized by dividing by 48 h, and terrain height $\hgt$ is normalized by subtracting the minimum value and dividing by 3000 m, ensuring that all values are in the range $[0,1]$. 

\section{Conditional Wasserstein Generative Adversarial Network (cWGAN)} \label{cWGAN}
To solve the target inverse problem of inferring fire arrival times  $\arrtime$ from measurements $\arrtimemeas$ and terrain height $\hgt$, we employ a cWGAN. The cWGAN comprises two subnetworks, a generator $\bm{g}$ and a critic $d$, which are given by the mappings $\bm{g}: \Omega_{\arrtimemeas} \times \Omega_{\hgt} \times \Omega_{\latent} \to \Omega_{\arrtime}$ and $d: \Omega_{\arrtime} \times \Omega_{\arrtimemeas} \times \Omega_{\hgt} \to \mathbb{R}$, respectively. Where $\latent \in \Omega_{\latent} \subset \mathbb{R}^{N_{\latent}}$ is a latent variable modeled using the random variable $\latentRV$ sampled from a multivariate Gaussian distribution $\latentdist$. Once trained, given a measurement-terrain pair $(\arrtimemeas,\hgt)$, the generator $\bm{g}$ produces samples $\arrtimegen=\bm{g}(\arrtimemeas,\hgt,\latent), \latent \sim \latentdist$ from the learned conditional distribution $\condgen(\arrtime|\arrtimemeas,\hgt)$. 

Training of the cWGAN requires that the Wasserstein-1 distance between the learned distribution $\condgen(\arrtime|\arrtimemeas,\hgt)$ and the true conditional distribution $\cond(\arrtime|\arrtimemeas,\hgt)$ be minimized. To accomplish this, a min-max problem is solved using the objective function 
\begin{equation} \label{obj_fun}
    \mathcal{L}(d,\bm{g}) = \mathop{\mathbb{E}}_{\substack{(\arrtime,\arrtimemeas,\hgt) \sim \joint}} [d(\arrtime,\arrtimemeas,\hgt)] - \mathop{\mathbb{E}}_{\substack{\arrtimegen \sim \condgen\\ (\arrtimemeas, \hgt) \sim P_{\arrtimemeasRV, \hgtRV}}} [d(\arrtimegen,\arrtimemeas,\hgt)],
\end{equation}
where the optimal generator and critic, denoted by $\bm{g}^*$ and $d^*$, respectively, are found as
\begin{equation} \label{minmax}
    (d^*,\bm{g}^*) = \mathop{\mathrm{arg \, min}}_{\bm{g}} \; \mathop{\mathrm{arg \, max}}_{d} \; \mathcal{L}(d,\bm{g}).
\end{equation}

The $\bm{g}^*$ found by solving the min-max problem given by Eq.~\ref{minmax} can then be used to approximate the true conditional distribution, assuming the critic is 1-Lipschitz in all of its arguments \citep{ray2023solution}. Selecting a generator with a sufficiently large number of learnable parameters, for any continuous, bounded function $\ell(\arrtime)$ defined on $\Omega_\arrtime$ and for $\epsilon>0$, it can be shown that 
\begin{equation} \label{conv}
    |\mathop{\mathbb{E}}_{\arrtime \sim \cond} [\ell(\arrtime)]  - \mathop{\mathbb{E}}_{\arrtimegen \sim P_{\bm{T}|\bm{\bar{T}}, \hgtRV}^{\bm{g}^*}} [\ell(\arrtimegen)] | < \epsilon. 
\end{equation}

Therefore, the trained generator may be used to approximate statistics from the true conditional distribution as 
\begin{equation} \label{stats}
    \mathop{\mathbb{E}}_{\arrtime \sim \cond} [\ell(\arrtime)] \approx \frac{1}{K} \sum\limits_{i=1}^K \ell (\bm{g}^*(\arrtimemeas,\hgt,\latent^{(i)})), \quad \latent^{(i)} \sim \latentdist,
\end{equation}
where by setting $\ell(\arrtime)=\arrtime$ the pixel-wise mean prediction may be computed, and by setting $\ell(\arrtime)=(\arrtime-\mathbb{E}[\arrtime])^2$ the pixel-wise variance may be computed. 

\subsection{Architecture}
\begin{figure}
    \centering
    \includegraphics[width=0.9\linewidth]{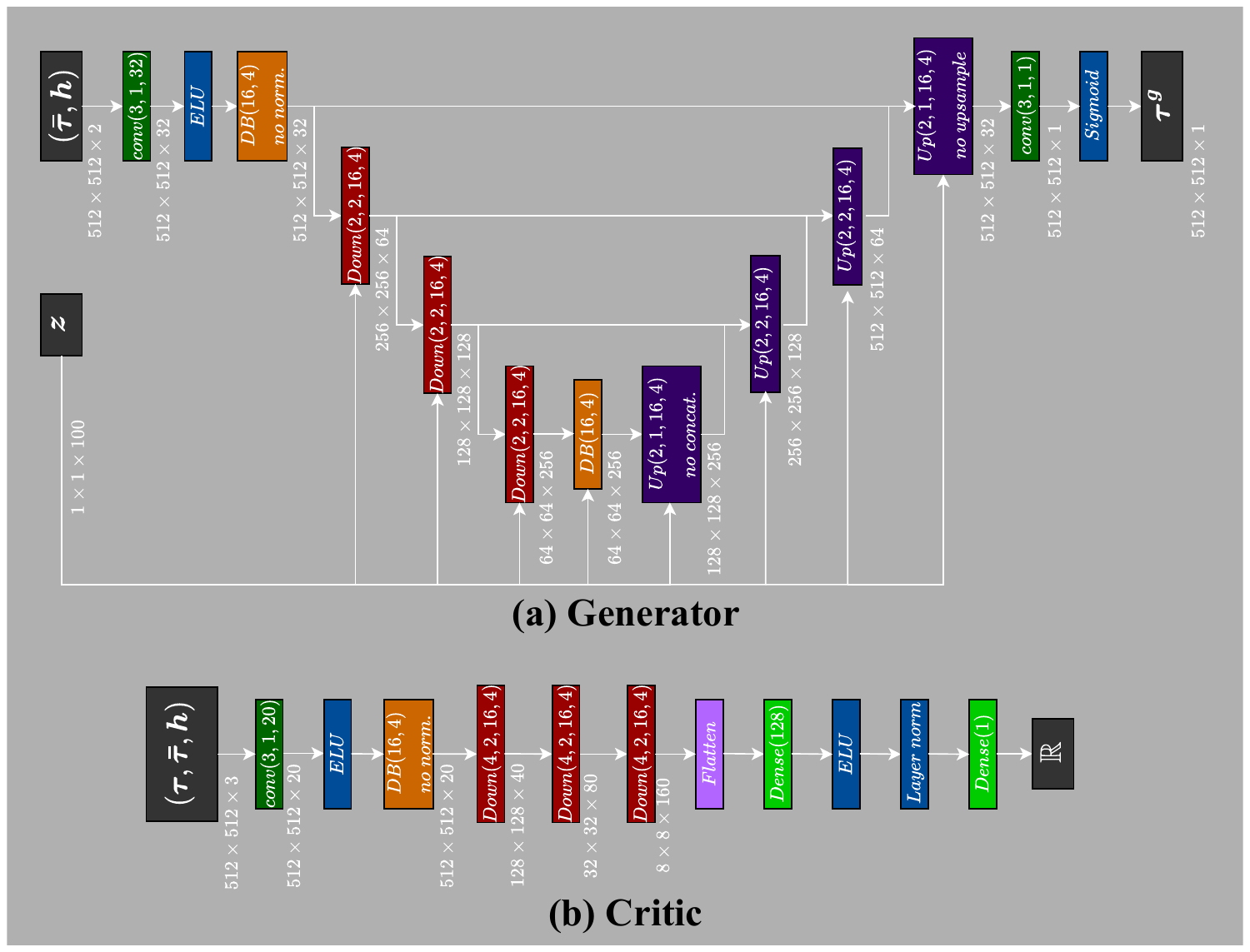}
    \caption{Network architecture for (a) generator and (b) critic of cWGAN.}
    \label{fig:generator_and_critic_architectures}
\end{figure}

Network architectures used for the generator and critic subnetworks are shown in Fig.~\ref{fig:generator_and_critic_architectures} and follow that of \cite{shaddy2024generative}, with the exception of an additional input channel to both to accommodate terrain conditioning. The generator uses a U-Net architecture which takes as input the measurement $\arrtimemeas$ and terrain height $\hgt$, along with the latent vector $\latent$. The generator is composed of dense blocks, down-sampling blocks, and up-sampling blocks, and uses conditional instance normalization (CIN) to inject latent information at multiple scales, overcoming issues of mode collapse \citep{dumoulin2016learned,adler2018deep}. The critic architecture is comprised of dense blocks, down-sampling blocks, and fully connected layers, and takes as input tuples of $(\arrtime,\arrtimemeas,\hgt)$ to produce a real-valued scalar. 

Dense blocks follow the work of \cite{huang2017densely} and are denoted as $DB(k,n)$, where $n$ is the number of sub-blocks and $k$ is the number of output features from intermediate sub-blocks. Down-sampling blocks consist of a convolutional layer, followed by ELU activation, average pooling, and a dense block, and are denoted as $Down(p,q,k,n)$, where $p$ denotes the factor by which images are coarsened from average pooling, $q$ denotes the factor by which channels are increased by convolutions, and $k$ and $n$ are the parameters of the dense block. Up-sampling blocks are denoted as $Up(p,q,k,n)$ and consist of concatenation for skip connections, convolutions, ELU activation, interpolation, and a dense block, where again $p$ indicates the factor by which images are refined by interpolation, $q$ is the factor by which channels are reduced, and $k$ and $n$ are parameters for the dense block. 

\subsection{Training}
The cWGAN was implemented in PyTorch and trained using 2 A100 GPUs for a total of 160 epochs. A batch size of 16 was used, along with the Adam optimizer with a learning rate of 0.001, $\beta_1$ of $0.5$, $\beta_2$ of $0.9$, and weight decay of 1e-7. Training progression was tracked using a mismatch term equal to the mean Frobenius norm of the difference between the true fire arrival time $\arrtime$ from the training set and generated fire arrival time samples $\arrtimegen$, for a given set of inputs $(\arrtimemeas,\hgt)$.

\section{Results} \label{results}
In the following section we apply the cWGAN-based approach for fire arrival time reconstruction to naturally occurring wildfires taking place between 2019 and 2022 in California and Washington. We first discuss the wildfire incidents considered, after which we present the GOES-derived ignition times for these cases and the VIIRS AF data utilized. Following this, the fire arrival time reconstructions are presented and validated using high resolution fire extent perimeters, along with a discussion of prediction uncertainty. Lastly, the influence of terrain conditioning on model performance is examined.

\subsection{Wildfire incidents}
The selection of wildfire incidents for validating the cWGAN-based approach for estimating fire arrival times is based on two primary factors. First, the extent of the wildfire after the initial 48 hours of burning should be contained within the domain size of 12.8 km by 12.8 km considered in the problem setup. Second, to allow for the validation of generated fire arrival time predictions, fire extent perimeters must be available within the first 48 hours of the fire. For this purpose high resolution infrared (IR) fire extent perimeters from the National Infrared Operations (NIROPS) program are utilized \citep{greenfield2003phoenix}. These perimeters are collected via IR sensors mounted aboard aircraft which fly over wildfires and map their extent at the time of measurement with resolutions of approximately 6.3 m/pixel. Following these considerations, five wildfires are selected which exhibit a variety of sizes, terrain conditions, and AF detection densities, namely the Bobcat, Tennant, Oak, Barnes, and Williams Flats fires. Additional information about these fires is provided in Table~\ref{tab:fires}.

\begin{table}[]
    \centering
    \begin{tabular}{c c c c c}
         Fire & \thead{Ignition \\ Date} & \thead{IR Measurement \\ Date} & \thead{IR N-S \\ Extent} & \thead{IR E-W \\ Extent} \\
         \hline
         Bobcat & 6 September 2020 & 8 September 2020 & 11.9 km & 7.8 km \\
         Tennant & 28 June 2021 & 30 June 2021 & 11.4 km & 7.7 km \\ 
         Oak & 22 July 2022 & 24 July 2022 & 10.2 km & 9.7 km \\
         Barnes & 7 September 2022 & 9 September 2022 & 3.6 km & 7.9 km \\
         \thead{Williams \\ Flats} & 2 August 2019 & 4 August 2019 & 6.5 km & 11.2 km 
    \end{tabular}
    \caption{Wildfires incidents utilized for model validation, along with their ignition date, date of first IR perimeter measurement, and approximate extent in the North-South and East-West directions.}
    \label{tab:fires}
\end{table}

\subsection{GOES ignition time estimates}
Ignition time estimates are gathered from the Geostationary Operational Environmental Satellite (GOES) Fire / Hot Spot Characterization product. The GOES fire product uses both visible and infrared spectral bands to locate actively burning fires with a spatial resolution of 2-4 km depending on the location being measured (coarser for locations farther from the measuring satellite) and a temporal resolution of 5 minutes \citep{schmidt2010goes}. The GOES fire product provides a geolocated fire mask which categorizes each pixel as either fire with an associated confidence level or as another non-fire category. GOES data has been collected using AWS, with GOES-West being utilized for the fires examined here, which occurred in the Western US, though GOES-East may also be utilized. 

To gather ignition time estimates from the GOES fire product, the following steps are followed:

\begin{enumerate}
    \item Collect GOES fire mask data from the Fire / Hot Spot Characterization product for time period of +/- 1 hour around approximate fire start time. 
    \item Select domain of size 12.8 km x 12.8 km centered on fire of interest.
    \item Find portion of collected fire masks intersecting chosen domain based on latitude and longitude coordinates of fire mask pixels. 
    \item Considering relevant portion of fire masks, find fire masks containing an active fire pixel (excluding low or nominal confidence pixels). 
    \item Use time of earliest detection for domain of interest as ignition time estimate. 
    \item If no GOES fire detection is found, collect additional data for larger time period around approximate start time and repeat process until a detection is found.
\end{enumerate}

Following this set of steps, an ignition time estimate is obtained for each fire examined, which is utilized to construct the measurement input to our network. Table~\ref{tab:ign_times} provides the ignition time estimates derived from GOES fire detections for the five fires examined here. 

\begin{table}[]
    \centering
    \begin{tabular}{c | c c c c c}
        Fire & Bobcat & Tennant & Oak & Barnes & \thead{Williams \\ Flats} \\
        \hline
        \thead{GOES Ignition \\ Time Estimate} & 1916 & 2321 & 2126 & 2351 & 1446 \\
    \end{tabular}
    \caption{Ignition time estimates (UTC) derived from GOES Fire / Hot Spot Characterization data product for validation fires.}
    \label{tab:ign_times}
\end{table}

\subsection{VIIRS active fire measurements}
Active fire measurements are obtained from the Collection 2 Visible Infrared Imaging Radiometer Suite (VIIRS) 375 m Active Fire (AF) Product utilizing the VIIRS system flown aboard the polar-oribiting Suomi-NPP and NOAA-20 satellites, which provide data products VNP14IMG and VJ114IMG, respectively. The VIIRS AF product uses infrared imaging to determine the location of actively burning fires during a 6 minute measurement period on a grid with 375 m resolution \citep{schroeder2024collection}. The final data product provides the latitude and longitude coordinates for each AF detection pixel, along with a corresponding confidence level of high, nominal, or low. Additionally, a categorical fire mask is provided which provides classifications such as high confidence fire, water, clouds, etc. for each 375 m pixel on a unique geolocated grid for each measurement file, where the measurement grid is dictated by the satellites position during measurement. Measurements are provided anywhere from 2 to 4 times a day per satellite, depending on the geographic location of interest relative to the orbit of the satellites. VIIRS AF data is collected from NASA LAADS DAAC based on the time period of interest and approximate fire location. 

\begin{figure}
    \centering
    \includegraphics[width=0.62\linewidth]{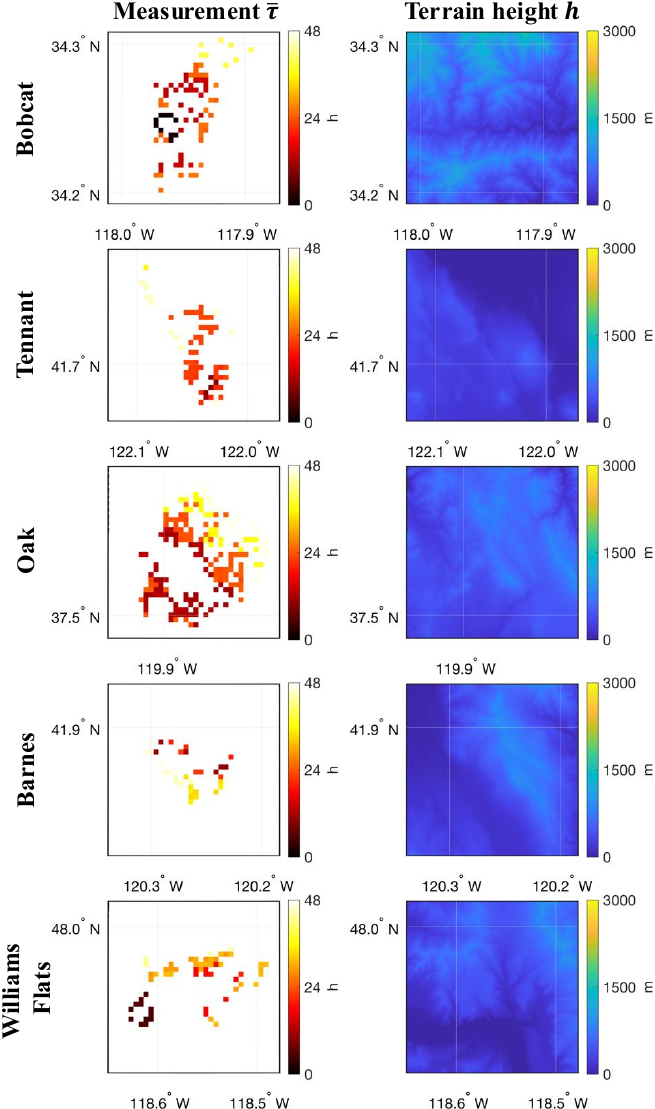}
    \caption{Inputs to the cWGAN model for the Bobcat, Tennant, Oak, Barnes, and Williams Flats fires. The inputs include  measurements based on VIIRS high confidence detections with GOES ignition times and terrain height maps.}
    \label{fig:meas_and_terr}
\end{figure}

Following collection of VIIRS AF data for a fire of interest, a geolocated discretized grid of size 34 x 34 with 375 m resolution centered over the fire of interest and co-located with the domain used for determining ignition times from GOES is created. Based on the latitude and longitude coordinates of high confidence AF detections, pixels in the discretized grid are assigned arrival time values based on the VIIRS measurement times and the ignition time estimate obtained from GOES (i.e., they are assigned a value equal to the VIIRS measurement time minus the GOES-derived ignition time, to provide an approximate arrival time value). Thereafter, unassigned pixels in the discretized grid are assigned a value of 48 h, following what is done for measurements in the training set. The measurement image is then resampled to the 25 m resolution used by the cWGAN, and padded to return to a total domain size of 12.8 km x 12.8 km. Fig.~\ref{fig:meas_and_terr} depicts the final measurements constructed for the fires examined here, along with their corresponding terrain height maps.

\subsection{Fire Arrival time predictions}
\begin{figure}
    \centering
    \includegraphics[width=0.62\linewidth]{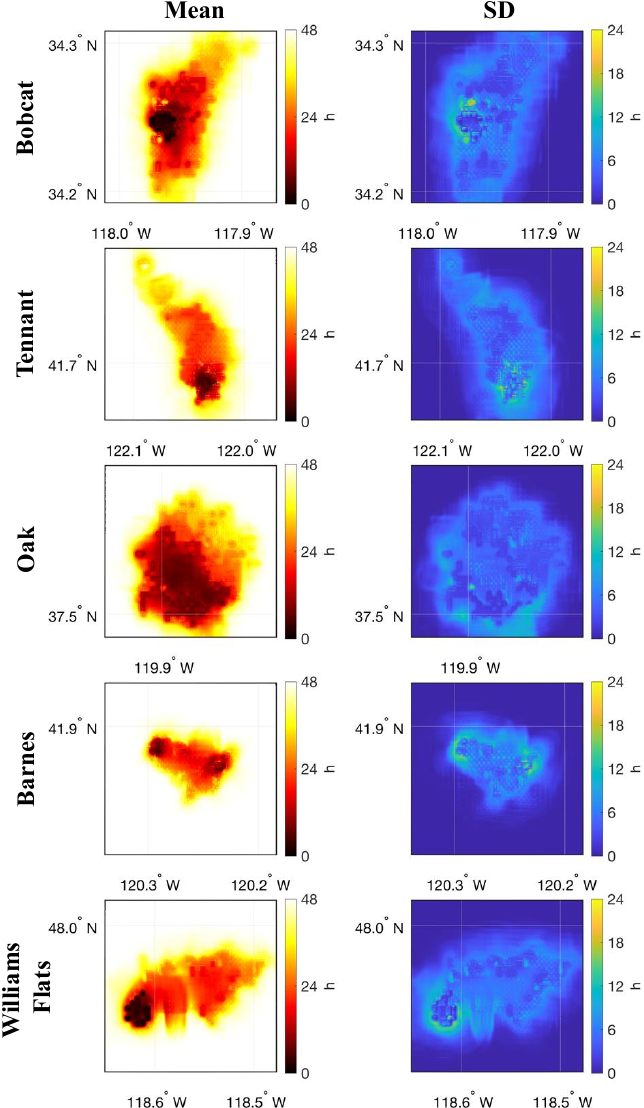}
    \caption{Mean fire arrival time predictions (left column) and standard deviation of fire arrival time predictions (right column) computed using 500 generated fire arrival time instances for the Bobcat, Tennant, Oak, Barnes, and Williams Flats fires.}
    \label{fig:mean_and_SD}
\end{figure}

The arrival time measurement and terrain height maps for the Bobcat, Tennant, Oak, Barnes, and Williams Flats fires are then used as input to the trained generator of the cWGAN model. For each measurement-terrain pair, 500 corresponding fire arrival time map instances $\arrtimegen$ are generated from the learned conditional distribution $\condgen$ by sampling the latent vector $\latent$ from its distribution $\latentdist$ and passing these through the generator. Generated samples are used to compute the pixel-wise mean prediction for each fire following Eq.~\ref{stats}, providing the best estimate of fire progression. The pixel-wise standard deviation is also computed across generated samples to provide an estimate of prediction uncertainty. Fig.~\ref{fig:mean_and_SD} shows the resulting mean fire arrival time predictions and the standard deviation of generated fire arrival times for the Bobcat, Tennant, Oak, Barnes, and Williams Flats fires.

From Fig.~\ref{fig:mean_and_SD} we see that the mean fire arrival time smoothly interpolates the AF detections provided as input, resulting in arrival time maps which may be used to generate a progression of fire perimeters consistent with the measured behavior. We also observe that even in cases where there is an absence of measurements for a large potion of the fire, such as the center of the Oak fire or the southern portion of the Williams Flats fire, the predicted fire progression captures fire growth through these regions. We note that arrival time predictions in locations corresponding to measurements are strongly influenced by provided measurement times, resulting in a lower standard deviation in these measurement rich regions and a higher standard deviation in measurement sparse regions. We see also that regions corresponding to a sharp gradient of arrival times in the measurement, such as the western portions of the Bobcat and Williams Flats fires, correspond to a higher uncertainty, as evidenced by the elevated standard deviation. This increased standard deviation appears to stem from the abrupt difference in measured arrival times between neighboring detections, which implies an abrupt and non-smooth fire progression. 

\begin{figure}
    \centering
    \includegraphics[width=\linewidth]{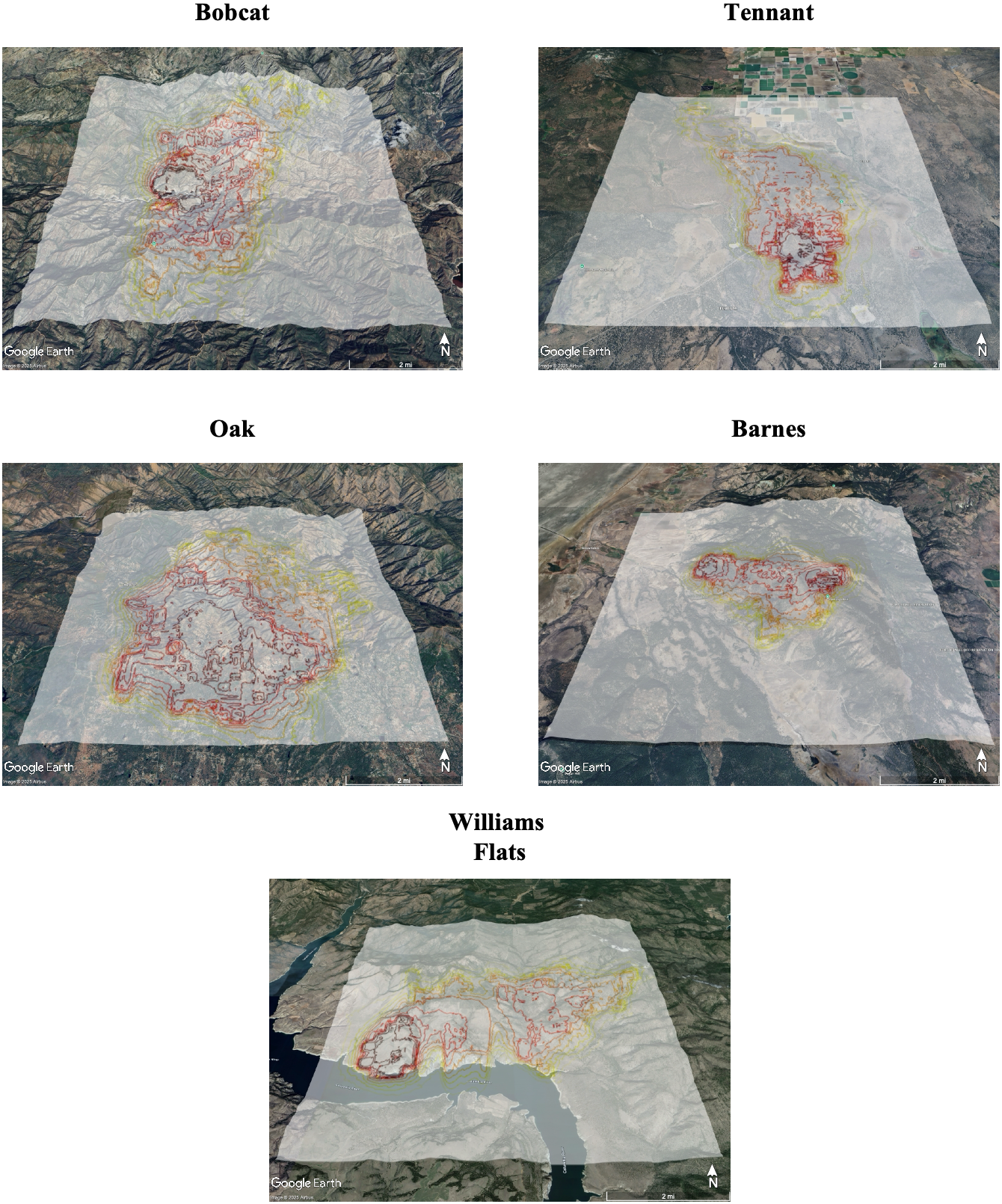}
    \caption{Contour maps of mean fire arrival time predictions for Bobcat, Tennant, Oak, Barnes, and Williams Flats fires displayed in Google Earth, with contours at 4 hour intervals.}
    \label{fig:arr_time_contours_GE}
\end{figure}

Using geolocation information for predicted fire arrival times, we can also visualize fire progression relative to geographical features visible through satellite imagery. Fig.~\ref{fig:arr_time_contours_GE} displays contour maps of the mean fire arrival time predictions generated for the fires of interest, shown in Google Earth, with contours at 4 hour intervals. These images provide a sense for how the fire front expanded over the course of each fire, displaying a sequence of predicted perimeters, and from these we can gather intuition about the behavior of each fire as it relates to terrain, vegetation, and other visible geographical features. Visualizations such as these may also prove useful for fire managers, where obtaining a measure of a fire's current extent could help decision making strategies. 

\subsection{Prediction validation}
To validate predicted fire arrival time maps, we compare estimated fire perimeters against available IR fire extent perimeters from NIROPS. Predicted fire perimeters are determined from the mean fire arrival time maps by taking contours corresponding to the times IR perimeters were measured. Based on the geolocation information for the fire arrival time maps, estimated perimeters are geolocated and directly comparable to measured IR perimeters. Based on the intersection of the estimated and measured perimeters three areas are determined: 1) area of agreement, 2) false negative area, and 3) false positive area, where areas of agreement account for areas where both the predicted and measured perimeters indicate fire presence, false negative areas account for regions where the measured perimeter indicates fire presence but the predicted perimeter does not, and false positive areas account for regions where fire is predicted but not present in the measured perimeter. 

\begin{figure}
    \centering
    \includegraphics[width=0.8\linewidth]{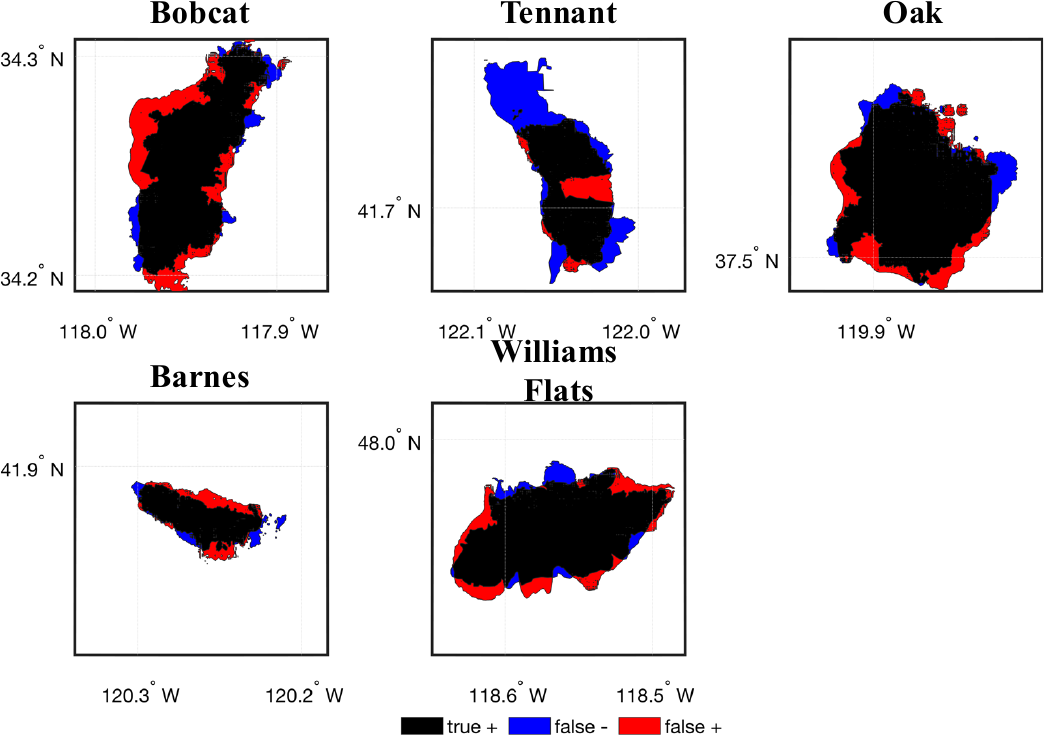}
    \caption{Comparison of predicted fire perimeters against measured high-resolution IR fire extent perimeters from NIROPS, with areas of agreement colored in black, false negative areas colored in blue, and false positive areas colored in red.}
    \label{fig:area_of_agreement}
\end{figure}

Fig.~\ref{fig:area_of_agreement} displays predicted perimeters for the Bobcat, Tennant, Oak, Barnes, and Williams Flats fires along with measured IR perimeters, indicating the three areas of interest. In this figure we observe that in most cases the cWGAN-based approach appears to capture the overall shape and extent of the wildfires at the time of IR perimeter measurement, with the exception of some finer details around the periphery which are not fully captured. We believe that these fine scale details likely stem from small-scale effects arising from interactions between weather, terrain, and fuel during fire spread, which the cWGAN model has no knowledge of beyond the limited information embedded in the measurement and terrain inputs. In the presence of higher spatial and temporal resolution measurement data it may be possible to capture more of these details, however even in this case the stochasticity of fire spread remains as a limiting factor. Examining comparisons between the predicted and measured perimeters further, we see that in most cases overall false positive area and false negative area are fairly equal in size, indicating that the model does not display an obvious bias. 

In the case of the Tennant fire the model under-predicts growth in the north and incorrectly predicts burning in the central region. We believe the northern false negative region arises from the large temporal gap in measurements, which jumped from roughly 22 hours after ignition to 45 hours after ignition, with the exception of a single detection around 36 hours in the northern-most position for this set of measurements. This temporal gap means that very little information about fire progression during the second day of growth was provided, resulting in the northern progression being predicted later than what is indicated by the IR perimeter. From Figure \ref{fig:mean_and_SD} we observe that the standard deviation in this region is slightly elevated relative to the center-most correctly predicted region, indicating higher uncertainty. Regarding the false positive growth in the central region of the fire, we believe that the growth pattern for the Tennant fire is somewhat unique relative to other fires examined here, for instance the Williams Flats fire had a similar region with no detections, however in that case fire did spread through the region and was correctly predicted by the cWGAN approach. 

We compare predicted and measured perimeters quantitatively by evaluating the S\o rensen-Dice coefficient (SC), the Probability of Detection (POD), and the False Alarm Ratio (FAR). SC is a measure of how well the two perimeters agree, POD is a measure of how well the prediction captures true fire growth, and FAR is a measure of false growth in predictions. These perimeters are defined as
\begin{equation}
    SC=\frac{2A}{2A+B+C} \quad,\quad POD=\frac{A}{A+B} \quad,\quad FAR=\frac{C}{A+C},
\end{equation}
where $A$ is the area of agreement between the two perimeters, $B$ is the false negative area, and $C$ is the false positive area. For all three metrics values range between 0 and 1, with the optimal value for the SC and POD being 1 and the optimal value for the FAR being 0. 

Table~\ref{tab:SC_POD_FAR} contains the SC, POD, and FAR values for the fires evaluated here. We observe that for all cases except the Tennant fire, the SC values are above 0.80 which indicates that the predicted fire extent agrees well with the measured IR perimeters. Additionally, POD values for Bobcat, Oak, and Williams Flats are above 0.90, indicating that in these cases the model does an excellent job of capturing the total fire growth present in the measured perimeters. In all cases the FAR values are less than 0.30, indicating a low tendency to falsely predict fire where fire is not present. Overall these metrics indicate that the model reasonably captures the wildfire extent at a given time, while balancing trade-offs between over and under predicting.

\begin{table}[]
    \centering
    \begin{tabular}{c|c c c}
        Fires & SC & POD & FAR \\
        \hline
        Bobcat & 0.82 & 0.91 & 0.26 \\
        Tennant & 0.67 & 0.57 & 0.17 \\
        Oak & 0.87 & 0.90 & 0.16 \\
        Barnes & 0.81 & 0.87 & 0.25 \\
        \thead{Williams \\ Flats} & 0.88 & 0.92 & 0.16
    \end{tabular}
    \caption{S\o rensen-Dice coefficient (SC), Probability of Detection (POD), and False Alarm Ration (FAR) values for predicted fire perimeters relative to high-resolution measured IR perimeters from NIROPS for the Bobcat, Tennant, Oak, Barnes, and Williams Flats fires.}
    \label{tab:SC_POD_FAR}
\end{table}

\subsection{Evaluation of terrain influence on arrival time reconstruction}
\begin{figure}
    \centering
    \includegraphics[width=0.52\linewidth]{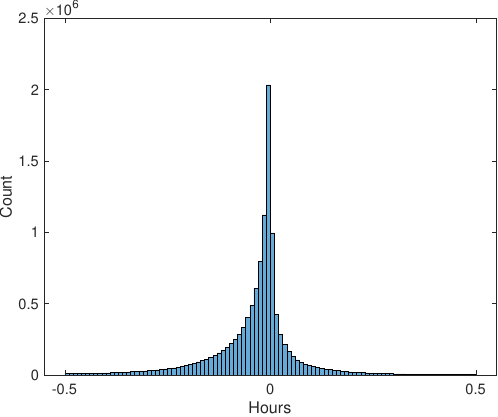}
    \caption{Histogram of pixel-wise difference in mean predicted fire arrival time maps conditioned on correct terrain and their counterparts conditioned on a flat terrain. The data was generated for 200 wildfires selected from a validation set.}
    \label{fig:hist}
\end{figure}

To evaluate the influence of terrain height conditioning on fire arrival time estimates, additional tests are undertaken. We run tests where, for a given measurement, fire arrival time predictions are conditioned on both the correct terrain height information and a constant zero-valued non-varying terrain height. For 200 measurement samples from the validation dataset, we generate 500 fire arrival time maps both for the correct terrain height data and for flat terrain. Thereafter, we compute the mean arrival time predictions for the two cases. Finally, we compute the pixel-wise difference in arrival times by subtracting the mean fire arrival time map for the case with the correct terrain information from the mean fire arrival time map for the case with flat terrain. This is done for each of the 200 samples considered from the validation set and a histogram of the difference is plotted in Figure ~\ref{fig:hist}. In creating this histogram pixels where the arrival time exceeds 47 hours in both are excluded so as to remove the effect of any unburned regions. 

From this figure we observe that in most instances the difference in predicted arrival times are very small, with more than $97 \%$ of pixels having a difference of less than 30 minutes. This indicates that in most cases, estimated arrival times have little variation resulting from conditioning on the terrain map. This suggests that the prediction is most strongly influenced by the arrival time measurement data, and once this data has been assimilated the effect of the terrain map is negligible. We do find that there is a minor skew towards negative values in the histogram, apparent both visually and also demonstrated by the mean difference value of -3 minutes for the histogram, indicating that there is a tendency for predictions conditioned on uniform terrain to be slightly lower in value than those conditioned on the true terrain map.

\section{Conclusion and outlook} \label{conclusion}
In this work we have developed an approach for estimating wildfire progression from satellite measurements and terrain height data using a conditional Wasserstein Generative Adversarial Network (cWGAN) trained on simulations of historic wildfires. The trained cWGAN generates samples from the conditional distribution of fire arrival times given satellite active wildfire detections, augmented with ignition time estimates from geostationary satellites, and terrain height data. Owing to the probabilistic framework of the approach, uncertainty information may also be gathered by computing the standard deviation of generated samples, and further, a best guess prediction is found from the mean of generated samples. Once obtained, reconstructed fire arrival time maps may be used to initialize forecasts from coupled atmosphere-wildfire models.

The training data is generated from WRF-SFIRE wildfire spread simulations of the initial 48 hours of 140 wildfires occurring in CONUS during 2023. Artificial satellite measurement data for training is obtained by applying an observation operator to the simulated fire arrival time maps, producing measurements that are consistent with true satellite data. By training with solutions from WRF-SFIRE, the wildfire spread physics of this model are incorporated into fire arrival time estimates produced by the trained cWGAN. 

When applied to real fires, active fire detections from the polar-orbiting VIIRS measurement system were utilized, along with ignition time estimates obtained from GOES data. The trained cWGAN was applied to wildfires occurring in California and Washington, and results were evaluated against high-resolution IR wildfire extent perimeters provided through the NIROPS program. Predicted wildfire perimeters were compared to IR perimeters, and a mean S\o rensen-Dice coefficient of 0.81 (0.85 when excluding the Tennant fire) was found across 5 wildfire instances, along with an average Probability of Detection of 0.83, and an average False Alarm Ratio of 0.20. Of the five fires examined, predictions for the Tennant fire exhibited the largest discrepancy relative to the measured IR perimeter, which appeared to result from insufficient AF detections which led to a temporal gap of roughly 20 hours in measurement availability. 

The influence of terrain height on fire arrival time predictions was also explored. It was observed that terrain height information included as a conditional input to the cWGAN ultimately had very little influence over generated predictions. For 200 test cases the difference between the mean fire arrival time prediction conditioned on true terrain maps and the mean prediction conditioned on a flat terrain was examined, and it was observed that $97\%$ of predicted fire arrival time values (excluding unburned pixels) had a difference of less than 30 minutes. It was concluded that conditioning fire arrival time predictions on terrain height data brought very little additional information beyond what was provided by the satellite measurements. This implies that in future studies terrain may be excluded in preference of other fields like the fuel map that are likely to be more informative.

\clearpage
\acknowledgments

The NOAA Bipartisan Infrastructure Law project NA22OAR4050672I supported creation of the forecasts. The NASA Disasters project 80NSSC19K1091 supported the high-performance computing for running the forecasts. The NASA FireSense project 80NSSC23K1344 supported development and evaluation of the machine learning approach. The authors acknowledge the Center for Advanced Research Computing (CARC) at the University of Southern California, USA for providing computing resources that have contributed to the research results reported within this publication.

%
%

%






%



\bibliographystyle{ametsocV6}
\bibliography{references}

\begin{thebibliography}{24}
\providecommand{\natexlab}[1]{#1}
\providecommand{\url}[1]{\texttt{#1}}
\renewcommand{\UrlFont}{\rmfamily}
\providecommand{\urlprefix}{URL }
\expandafter\ifx\csname urlstyle\endcsname\relax
  \providecommand{\doi}[1]{https://doi.org/\discretionary{}{}{}#1}\else
  \providecommand{\doi}{https://doi.org/\discretionary{}{}{}\begingroup \urlstyle{rm}\Url}\fi
\providecommand{\eprint}[2][]{\url{#2}}

\bibitem[{Adler and {\"O}ktem(2018)Adler, and {\"O}ktem}]{adler2018deep}
Adler, J., and O.~{\"O}ktem, 2018: Deep bayesian inversion. \textit{arXiv preprint arXiv:1811.05910}, \doi{10.48550/arXiv.1811.05910}.

\bibitem[{Albini(1976)}]{albini1976estimating}
Albini, F.~A., 1976: \textit{Estimating wildfire behavior and effects}. General technical report, INT-30. USDA Forest Service, Intermountain Forest and Range Experiment Station.

\bibitem[{Bakhshaii and Johnson(2019)Bakhshaii, and Johnson}]{bakhshaii2019review}
Bakhshaii, A., and E.~A. Johnson, 2019: A review of a new generation of wildfire--atmosphere modeling. \textit{Can.\ J.\ For.\ Res.}, \textbf{49~(6)}, 565--574, \doi{10.1139/cjfr-2018-0138}.

\bibitem[{Burke et~al.(2021)Burke, Driscoll, Heft-Neal, Xue, Burney,, and Wara}]{burke2021changing}
Burke, M., A.~Driscoll, S.~Heft-Neal, J.~Xue, J.~Burney, and M.~Wara, 2021: The changing risk and burden of wildfire in the united states. \textit{Proceedings of the National Academy of Sciences}, \textbf{118~(2)}, e2011048\,118.

\bibitem[{Danielle(2025)}]{Danielle_2025}
Danielle, M., 2025: Accuweather estimates more than \$250 billion in damages and economic loss from la wildfires. AccuWeather, \urlprefix\url{https://www.accuweather.com/en/weather-news/accuweather-estimates-more-than-250-billion-in-damages-and-economic-loss-from-la-wildfires/1733821}.

\bibitem[{Dennison et~al.(2014)Dennison, Brewer, Arnold,, and Moritz}]{dennison2014large}
Dennison, P.~E., S.~C. Brewer, J.~D. Arnold, and M.~A. Moritz, 2014: Large wildfire trends in the western {United States}, 1984--2011. \textit{Geophysical Research Letters}, \textbf{41~(8)}, 2928--2933, \doi{10.1002/2014GL059576}.

\bibitem[{Dumoulin et~al.(2016)Dumoulin, Shlens,, and Kudlur}]{dumoulin2016learned}
Dumoulin, V., J.~Shlens, and M.~Kudlur, 2016: A learned representation for artistic style. \textit{arXiv preprint arXiv:1610.07629}, \doi{10.48550/arXiv.1610.07629}.

\bibitem[{Farguell et~al.(2021)Farguell, Mandel, Haley, Mallia, Kochanski,, and Hilburn}]{farguell2021machine}
Farguell, A., J.~Mandel, J.~Haley, D.~V. Mallia, A.~Kochanski, and K.~Hilburn, 2021: Machine learning estimation of fire arrival time from {Level-2 Active Fires} satellite data. \textit{Remote Sens.}, \textbf{13~(11)}, 2203, \doi{10.3390/rs13112203}.

\bibitem[{Flannigan et~al.(2013)Flannigan, Cantin, De~Groot, Wotton, Newbery,, and Gowman}]{flannigan2013global}
Flannigan, M., A.~S. Cantin, W.~J. De~Groot, M.~Wotton, A.~Newbery, and L.~M. Gowman, 2013: Global wildland fire season severity in the 21st century. \textit{Forest Ecology and Management}, \textbf{294}, 54--61.

\bibitem[{Greenfield et~al.(2003)Greenfield, Smith,, and Chamberlain}]{greenfield2003phoenix}
Greenfield, P.~H., W.~Smith, and D.~C. Chamberlain, 2003: Phoenix-the new forest service airborne infrared fire detection and mapping system. \textit{2nd Int. Wildland Fire Ecology and Fire Management Congress and the 5th Symposium on Fire and Forest Meteorology}.

\bibitem[{Huang et~al.(2017)Huang, Liu, Van Der~Maaten,, and Weinberger}]{huang2017densely}
Huang, G., Z.~Liu, L.~Van Der~Maaten, and K.~Q. Weinberger, 2017: Densely connected convolutional networks. \textit{Proceedings of the IEEE conference on computer vision and pattern recognition}, 4700--4708, \doi{10.48550/arXiv.1608.06993}.

\bibitem[{Lareau and Clements(2016)Lareau, and Clements}]{lareau2016environmental}
Lareau, N.~P., and C.~B. Clements, 2016: Environmental controls on pyrocumulus and pyrocumulonimbus initiation and development. \textit{Atmospheric Chemistry and Physics}, \textbf{16~(6)}, 4005--4022.

\bibitem[{Lareau and Clements(2017)Lareau, and Clements}]{lareau2017mean}
Lareau, N.~P., and C.~B. Clements, 2017: The mean and turbulent properties of a wildfire convective plume. \textit{J.\ Appl.\ Meteor.\ Climatol.}, \textbf{56~(8)}, 2289--2299, \doi{10.1175/JAMC-D-16-0384.1}.

\bibitem[{Mandel et~al.(2011)Mandel, Beezley,, and Kochanski}]{mandel2011coupled}
Mandel, J., J.~Beezley, and A.~Kochanski, 2011: Coupled atmosphere-wildland fire modeling with wrf-fire version 3.3. \textit{Geoscientific Model Development Discussions}, \textbf{4~(1)}, 497--545.

\bibitem[{Mandel et~al.(2012)Mandel, Beezley, Kochanski, Kondratenko,, and Kim}]{mandel2012assimilation}
Mandel, J., J.~D. Beezley, A.~K. Kochanski, V.~Y. Kondratenko, and M.~Kim, 2012: Assimilation of perimeter data and coupling with fuel moisture in a wildland fire--atmosphere dddas. \textit{Procedia Computer Science}, \textbf{9}, 1100--1109, \doi{10.1016/j.procs.2012.04.119}.

\bibitem[{Mandel et~al.(2014)}]{mandel2014recent}
Mandel, J., and Coauthors, 2014: Recent advances and applications of {WRF--SFIRE}. \textit{Natural Hazards and Earth System Sciences}, \textbf{14~(10)}, 2829--2845, \doi{10.5194/nhess-14-2829-2014}.

\bibitem[{Pastor et~al.(2003)Pastor, Z{\'a}rate, Planas,, and Arnaldos}]{pastor2003mathematical}
Pastor, E., L.~Z{\'a}rate, E.~Planas, and J.~Arnaldos, 2003: Mathematical models and calculation systems for the study of wildland fire behaviour. \textit{Progress in Energy and Combustion Science}, \textbf{29~(2)}, 139--153.

\bibitem[{Ray et~al.(2023)Ray, Murgoitio-Esandi, Dasgupta,, and Oberai}]{ray2023solution}
Ray, D., J.~Murgoitio-Esandi, A.~Dasgupta, and A.~A. Oberai, 2023: Solution of physics-based inverse problems using conditional generative adversarial networks with full gradient penalty. \textit{arXiv preprint arXiv:2306.04895}, \doi{10.48550/arXiv.2306.04895}.

\bibitem[{Rothermel(1972)}]{rothermel1972mathematical}
Rothermel, R.~C., 1972: \textit{A mathematical model for predicting fire spread in wildland fuels}. Research Paper, INT-115. US Department of Agriculture, Intermountain Forest and Range Experiment Station.

\bibitem[{Schmidt et~al.(2010)Schmidt, Hoffman, Prins,, and Lindstrom}]{schmidt2010goes}
Schmidt, C., J.~Hoffman, E.~Prins, and S.~Lindstrom, 2010: Goes-r advanced baseline imager (abi) algorithm theoretical basis document for fire/hot spot characterization, version 2.0, noaa, silver spring, md. \textit{NOAA NESDIS, Cent. Satell. Appl. Res}.

\bibitem[{Schroeder et~al.(2024)Schroeder, Giglio,, and Hall}]{schroeder2024collection}
Schroeder, W., L.~Giglio, and J.~Hall, 2024: Collection 2 visible infrared imaging radiometer suite (viirs) 375-m active fire product user’s guide version 1.0.

\bibitem[{Shaddy et~al.(2024)}]{shaddy2024generative}
Shaddy, B., and Coauthors, 2024: Generative algorithms for fusion of physics-based wildfire spread models with satellite data for initializing wildfire forecasts. \textit{Artificial Intelligence for the Earth Systems}.

\bibitem[{Wang et~al.(2021)}]{wang2021economic}
Wang, D., and Coauthors, 2021: Economic footprint of {California} wildfires in 2018. \textit{Nat.\ Sustain.}, \textbf{4~(3)}, 252--260, \doi{10.1038/s41893-020-00646-7}.

\bibitem[{Westerling et~al.(2006)Westerling, Hidalgo, Cayan,, and Swetnam}]{westerling2006warming}
Westerling, A.~L., H.~G. Hidalgo, D.~R. Cayan, and T.~W. Swetnam, 2006: Warming and earlier spring increase western {US} forest wildfire activity. \textit{science}, \textbf{313~(5789)}, 940--943, \doi{10.1126/science.1128834}.

\end{thebibliography}

\end{document}